\documentclass{article}
\usepackage{authblk}
\usepackage{amssymb}
\usepackage[dvipdfmx]{graphicx}
\usepackage{textcomp}
\usepackage{xcolor}
\usepackage{amsmath}
\usepackage{mathtools}
\usepackage{amsfonts}
\usepackage{accents}
\usepackage{bm}
\usepackage{subfig}
\usepackage{algorithm}
\usepackage{algorithmic}
\usepackage{listings}
\usepackage{multirow}
\usepackage{flushend}
\usepackage{geometry}
\usepackage{url}

\newcommand\vecspace[1]{\mathbb{R}^{#1}}
\newlength{\dtildeheight}

\begin{document}

\title{\vspace*{6mm}
Automatic Source Code Summarization with Extended Tree-LSTM
}
\date{}

\author[1]{Yusuke Shido\thanks{shido@iip.ist.i.kyoto-u.ac.jp}}
\author[1]{Yasuaki Kobayashi}
\author[1]{Akihiro Yamamoto}
\affil[1]{Graduate School of Informatics, Kyoto University, Japan}
\author[2]{Atsushi Miyamoto}
\author[2]{Tadayuki Matsumura}
\affil[2]{Center for Exploratory Research, Hitachi, Ltd., Japan}

\maketitle

\begin{abstract}
Neural machine translation models are used to automatically generate a document from given source code since this can be regarded as a machine translation task.
Source code summarization is one of the components for automatic document generation, which generates a summary in natural language from given source code.
This suggests that techniques used in neural machine translation, such as Long Short-Term Memory (LSTM), can be used for source code summarization. 
However, there is a considerable difference between source code and natural language: Source code is essentially {\em structured}, having loops and conditional branching, etc.
Therefore, there is some obstacle to apply known machine translation models to source code.

Abstract syntax trees (ASTs) capture these structural properties and play an important role in recent machine learning studies on source code.
Tree-LSTM is proposed as a generalization of LSTMs for tree-structured data.
However, there is a critical issue when applying it to ASTs: It cannot handle a tree that contains nodes having an arbitrary number of children and their order simultaneously, which ASTs generally have such nodes.
To address this issue, we propose an extension of Tree-LSTM, which we call \emph{Multi-way Tree-LSTM} and apply it for source code summarization.
As a result of computational experiments, our proposal achieved better results when compared with several state-of-the-art techniques.
\end{abstract}

\section{Introduction}\label{sec:intro}
In developing and maintaining software, it is desirable that details about a program, such as its package dependencies and behavior, are appropriately commented in its source code files to enable readers to understand the program's usage and purpose.
Given this, software developers are strongly encouraged to document source code.
However, documentation is often inaccurate, misleading, or even omitted because it is costly to write accurate and effective documentation, leading to developers spending a lot of time reading the source code \cite{Xia2018}.
To address this issue, automatic document generation has been studied in many software engineering studies.
{\em Source code summarization} is an important component of automatic document generation, which generates a short natural language summary from the source code.

Recent studies on source code summarization showed that high quality comments can be automatically generated with deep neural networks trained on a large-scale corpus \cite{Iyer2016, Hu2018a}.
To generate a good summary, a machine learning model needs to learn the functionality of the source code and translates it into natural language sentences.
Since the structural properties of source code are of a different nature from those in natural language, that is, they have loops, conditional branching, etc., 
we should leverage such properties rather than sequential representations of source code.
In many programming languages, the source code can be parsed into a tree-structured representation called an \emph{abstract syntax tree} (AST), which enables us to use structural information of the source code.
Several studies have reported that the results various tasks related to source code were improved by utilizing ASTs. Such tasks include classifying source code \cite{Mou2014}, code clone detection \cite{White2016}, predicting of method name \cite{Alon2018} and source code summarization \cite{Hu2018a,Wan2018}, which is the focus of this paper.

Long Short-Term Memory (LSTM) networks
\cite{hochreiter1997long}
play an important role in neural machine translation. This network is suitable for sequential data such as natural language sentences. However, due to the structured nature of source code, it may not be applicable to the sequential representation of source code.
\begin{figure}[tbp]
\centerline{\includegraphics[width=0.4\textwidth]{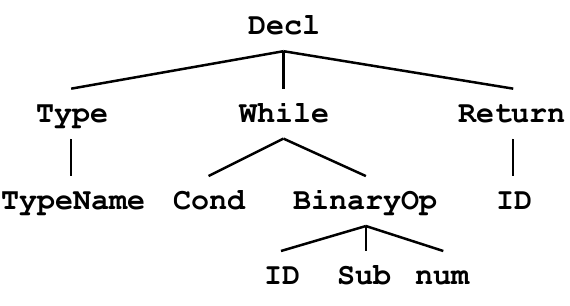}}
\caption{AST example. Each node has an arbitrary number of children, and the order is important in the program.}
\label{ast}
\end{figure}

Tree-LSTM \cite{tai2015}, originally proposed for predicting the semantic relatedness of two sentences and for sentiment classification, is a neural network architecture that handles tree-structured data, such as ASTs. It can be applied to other natural language processing (NLP) tasks (e.g. machine translation \cite{eriguchi2016}).
Tai {\it et al.} proposed two types of Tree-LSTM in their paper: The first type can handle trees in which each node has an arbitrary number of children, and the other type can handle the order of a fixed number of children at each node.
However, it is difficult to apply them to ASTs since ASTs have a node that has an arbitrary number of ordered children as in Figure~\ref{ast}.
In this research, we propose an extension of Tree-LSTM to solve this issue and use it as an encoder in our source code summarization model.

The contributions of this paper are shown below.
\begin{itemize}
\item We propose an extension of Tree-LSTM: The \emph{Multi-way Tree-LSTM} unit can handle a tree which contain a node having an arbitrary number of ordered children in ASTs.
\item We show that a tree-structured model with Multi-way Tree-LSTM, which can learn tree structures in ASTs directly, is more effective than a sequential model used for machine translation in NLP when applied to source code summarization.
\end{itemize}
To evaluate our model, we conducted computational experiments using a dataset consisting of pairs of a method and its documentation comment. Our experimental results show that our model is significantly better when compared with a state-of-the-art summarization model due to \cite{Hu2018a}, and some source code summaries generated by our model are more expressive than those in the original dataset.

\section{Background}\label{sec:back}
Source code summarization is related to machine translation.
Recently, Recurrent Neural Networks (RNNs) and LSTM are of a great importance in the NLP field.
In this section, we review some concepts and previous work related to our study.

\subsection{Recurrent Neural Networks}
\begin{figure}[tbp]
    \centering
    \subfloat[Standard RNN architecture and its unfolding\label{seq}]{%
        \centering
        \includegraphics[width=0.4\textwidth]{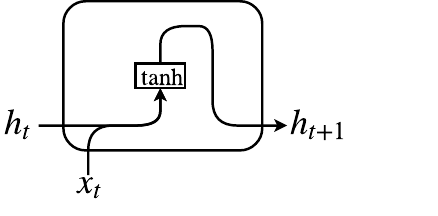}
        \raisebox{3mm}{\includegraphics[width=0.53\linewidth]{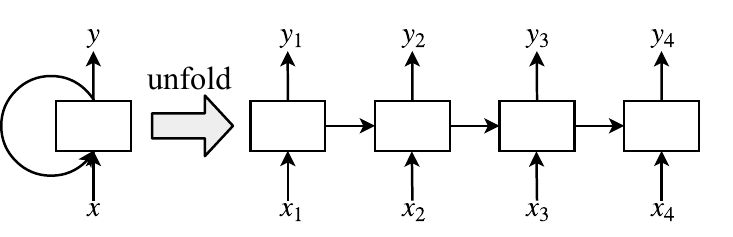}}
    }
    \\
    \subfloat[LSTM architecture\label{lstm}]{%
        \includegraphics[width=0.4\textwidth]{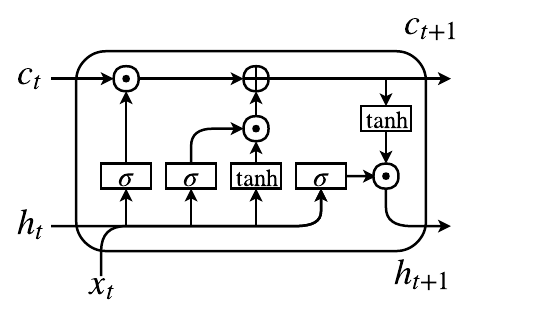}}
    \subfloat[Tree structured RNN architecture\label{tree}]{%
        \raisebox{3mm}{\includegraphics[width=0.4\textwidth]{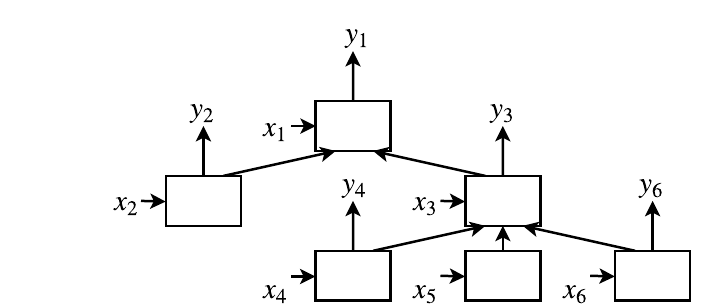}}}
\caption{Various recurrent neural networks.}
\end{figure}

RNNs have been frequently used in the NLP field.
Unlike feed-forward neural networks, RNNs take sequences of arbitrary lengths as input and generate sequences of the same length while updating their internal states as shown in Figure \ref{seq}.
Since sentences in natural languages can be seen as sequences of words, RNNs are well-suited to NLP.

The standard RNN receives a sequence of $d_1$-dimensional vectors $\bm{x}=(x_1, \ldots, x_n)$ as input and outputs a sequence of $d_2$-dimensional vectors while updating the hidden state $h_t$ at each time step $t$ as
	$h_t = \tanh (Wx_t + Uh_{t-1} + b)$,
where $x_t \in \vecspace{d_1}$ and $h_{t} \in \vecspace{d_2}$ are the input and hidden state vectors at time step $t$, respectively, $W \in \vecspace{d_2 \times d_1}$, $U \in \vecspace{d_2 \times d_2}$, and $b \in \vecspace{d_2}$ are model parameters.
Here, $\tanh$ denotes the hyperbolic tangent and is used as an activation function.

\subsubsection{Long Short-Term Memory (LSTM)}
Standard RNNs are not capable of learning ``long-term dependencies'', that is, they may not propagate information that appeared earlier in the input sequence later because of the vanishing and exploding gradient problems.
LSTM \cite{hochreiter1997long} has additional internal states, called {\em memory cells}, that do not suffer from the vanishing gradients and it controls what information will be propagated using {\em gates} as shown in Figure \ref{lstm}.
LSTM contains three independent gates. A {\em forget gate} discards irrelevant information from the memory cell. An {\em input gate} adds new information to the memory cell. An {\em output gate} computes the new hidden state.
With these structures, we can avoid vanishing gradients and train RNNs on long sequences, which can be used in various applications in the NLP field \cite{sundermeyer2012lstm}.
For each time step $t$, each unit in the LSTM can be computed by the following equations:
\begin{eqnarray*}{rCl}
    f_t &=& \sigma(W^{(f)} x_t + U^{(f)} h_{t-1} + b^{(f)}), \label{lstmf} \\
	u_t &=& \tanh (W^{(u)} x_t + U^{(u)} h_{t-1} + b^{(u)}), \label{lstmu} \\
	i_t &=& \sigma(W^{(i)} x_t + U^{(i)} h_{t-1} + b^{(i)}), \label{lstmi} \\
	o_t &=& \sigma(W^{(o)} x_t + U^{(o)} h_{t-1} + b^{(o)}), \label{lstmo} \\
	c_t &=& c_{t-1} \odot f_t + i_t \odot u_t, \label{lstmc} \\
	h_t &=& o_t \odot \tanh(c_t), \label{lstmh}
\end{eqnarray*}
where $f_t$, $i_t$ and $o_t$ denote the forget gate, the input gate, and the output gate for time step $t$, respectively, $\sigma$ denotes the sigmoid function, and $\odot$ denotes an element-wise product over matrices.
The model parameters $W^{(\ast)}$, $U^{(\ast)}$ and $b^{(\ast)}$ are matrices and vectors for $f_t$, $u_t$, $i_t$, and $o_t$.

\subsubsection{Tree-LSTMs}
We have seen that LSTM networks generate a sequence from an input sequence. Tai {\it et al.} \cite{tai2015} extended this type of network to generate a tree from an input tree, which they call Tree-LSTMs.
For each time step, standard LSTMs take an input vector and a single hidden state vector from the previous time step and propagate information from forward to backward. Tree-LSTMs can take multiple hidden states and propagate information from leaves to the root as shown in Figure \ref{tree}.
Tai {\it et al.} \cite{tai2015} proposed two kinds of Tree-LSTMs: \emph{Child-sum Tree-LSTM} and \emph{N-ary Tree-LSTM}.

\begin{figure}[tbp]
    \centering
    \subfloat[Child-sum Tree-LSTM \label{childsum}]{%
       \includegraphics[width=0.35\textwidth]{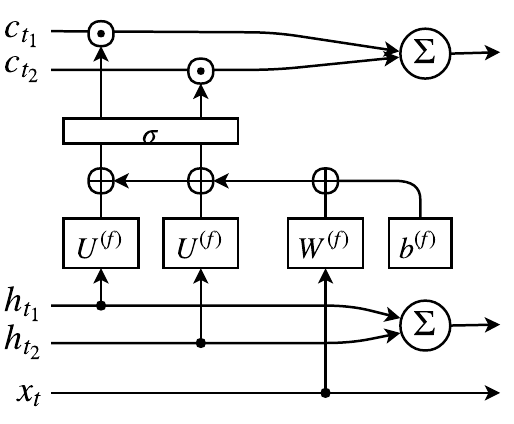}}
       \hspace{2cm}
    \subfloat[N-ary Tree-LSTM \label{nary}]{%
        \includegraphics[width=0.35\textwidth]{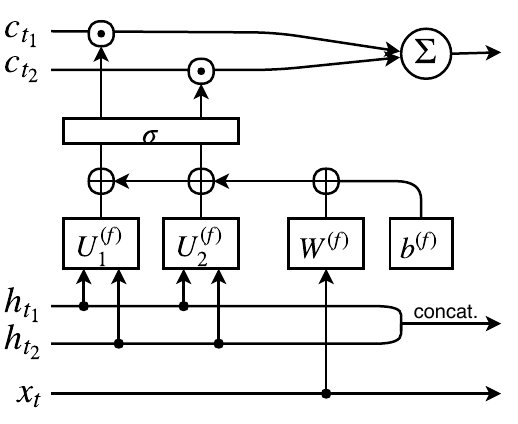}}
\caption{Forget gates in the Tree-LSTMs.}
\end{figure}

\bigskip
\noindent{\bf Child-sum Tree-LSTM}:
For an input vector $x_j$, we denote $C(j)$ as the children of $j$ and $n_j$ as the number of children $|C(j)|$.
In \emph{Child-sum Tree-LSTM}, the memory cell $c_j$ and the hidden state $h_j$ are computed as follows:
\begin{eqnarray}{rCl}
    \tilde{h_j} &=& \sum_{k \in C(j)} h_k, \label{csumsum} \\
	f_{j_k} &=& \sigma(W^{(f)} x_j + U^{(f)} h_k + b^{(f)}), \label{csumf} \\
	u_j &=& \tanh (W^{(u)} x_j + U^{(u)} \tilde{h_j} + b^{(u)}), \label{csu} \\
	i_j &=& \sigma(W^{(i)} x_j + U^{(i)} \tilde{h_j} + b^{(i)}), \label{csi} \\
	o_j &=& \sigma(W^{(o)} x_j + U^{(o)} \tilde{h_j} + b^{(o)}), \label{cso} \\
	c_j &=& \sum_{k \in C(j)}{c_k \odot f_{j_k}} + i_j \odot u_j, \label{csc} \\
	h_j &=& o_j \odot \tanh(c_j), \label{csh}
\end{eqnarray}
where $f_{t_k}$, $i_t$, and $o_t$ denote the forget gates, the input gate, and the output gate for time step $t$, respectively.
Note that the summation (\ref{csumsum}) of the hidden states $h_k$ of the children is given as input to \eqref{csu}, the input gate \eqref{csi} and the output gate \eqref{cso} and the same parameter $U^{(f)}$ is used for all the hidden states $h_k$ of children of $j$ in the forget gates \eqref{csumf}.
In the evaluating equation (\ref{csumf}), the parameters $U^{(f)} \in \vecspace{d_2 \times d_2}$ are shared for all children $C(j)$.
Therefore, the Child-sum Tree-LSTM can handle an arbitrary number of children.
As shown in Figure~\ref{childsum}, since the forget gate is independently computed for each child $k$, interactions among children are not taken into consideration when discarding information in the forget gate.
Furthermore, with the exception of the forget gate, the order of the children cannot be considered since information $h_k$ propagated from the children cannot be distinguished due to the summation (\ref{csumsum}).

\bigskip
\noindent{\bf N-ary Tree-LSTM}:
In \emph{N-ary Tree-LSTM}, the memory cell $c_j$ and the hidden state $h_j$ are computed as follows:
\begin{eqnarray}{rCl}
    \hat{h_j} &=& [h_{j_1};\cdots;h_{j_{n_{j}}}], \label{concath} \\
	f_{j_k} &=& \sigma(W^{(f)} x_j + U^{(f)}_k \hat{h_j} + b^{(f)}), \label{naf} \\
	u_j &=& \tanh (W^{(u)} x_j + U^{(u)} \hat{h_j} + b^{(u)}), \label{nau} \\
	i_j &=& \sigma(W^{(i)} x_j + U^{(i)} \hat{h_j} + b^{(i)}), \label{nai} \\
	o_j &=& \sigma(W^{(o)} x_j + U^{(o)} \hat{h_j} + b^{(o)}), \label{nao} \\
	c_j &=& \sum_{k \in C(j)}{c_k \odot f_{j_k}} + i_j \odot u_j, \label{nac} \\
	h_j &=& o_j \odot \tanh(c_j), \label{nah}
\end{eqnarray}
where $\hat{h_j} \in \vecspace{d_2 n_j}$ is the vector obtained by concatenating $n_j$ vectors $h_{j_1}, \ldots, h_{j_{n_{j}}}$.
Unlike Child-sum Tree-LSTM, parameters $U^{(f)}_k \in \vecspace{d_2 \times d_2 n_j}$ are not shared among the children and the concatenation (\ref{concath}) is used in \eqref{nau}, the input gate \eqref{nai}, and the output gate (\ref{nao}) instead of the summation (\ref{csumsum}).
As shown in Figure~\ref{nary}, interactions among children can be considered when discarding information in the forget gate since the forget gate is computed by the concatenation (\ref{concath}), and moreover, the children can be distinguished in (\ref{nau}), (\ref{nai}), and (\ref{nao}) due to the concatenation.
However, it is impossible to input trees containing nodes that have an arbitrary number of children because the size of parameter matrices must be fixed to compute the equations from \eqref{naf} to \eqref{nao}.

These Tree-LSTMs are not appropriate for ASTs of source code since they have nodes with an arbitrary number of children and their order is significant.
In previous studies (e.g. \cite{Wan2018}) of source code summarization, ASTs are converted into binary trees for applying the N-ary Tree-LSTM.

\subsection{Related Work}\label{sec:related}
Various methods for automatic source code summarization have been proposed.
There are several non-neural approaches: methods based on call relationships \cite{mcburney2014automatic} and topic modeling \cite{MovshovitzAttias2013}.
Oda {\it et al.} \cite{Oda2016} proposed a pseudocode generation method, which generates line-by-line comments from given source code.

Our focus is on neural network-based source code summarization.
In our approach, we train neural networks on a large-scale parallel corpus consisting of pairs of a method and its documentation comment.
This approach is frequently used in recent source code summarization studies.
Iyer {\it et al.} \cite{Iyer2016} proposed a neural source code summarization method based on an LSTM network with attention, called CODE-NN, and showed that this approach is promising for source code summarization as well as machine translation.
DeepCom \cite{Hu2018a} exploits the structural property of source code by means of ASTs. DeepCom is given an AST as a sequence obtained by traversing it and encodes the sequence with an LSTM encoder.
Note that the given AST is uniquely reconstructible from the encoded sequence they used.
ASTs are extensively used not only in code summarization studies but also in various software engineering studies \cite{White2016,Alon2018,Wei2017,Alsulami2017}.

\section{Proposed Approach}\label{sec:prop}
In this section, we propose an extension of Tree-LSTM and describe our code summarization framework.
\subsection{Multi-way Tree-LSTM}
As mentioned in Section~\ref{sec:back}, standard Tree-LSTMs proposed by Tai \textit{et al.} \cite{tai2015} cannot handle a node that has an arbitrary number of children and their order in ASTs simultaneously.
To overcome this difficulty, we develop an extension of Tree-LSTM, which we call \emph{Multi-way Tree-LSTM}.
The key to our extension is that we use LSTMs to encode the information of ordered children.
This idea enables us not only to handle an arbitrary number of ordered children but also to consider some interactions among children, which can take advantage in both Child-sum and N-ary Tree-LSTMs.

\begin{figure*}[tbp]
\centerline{\includegraphics[width=.9\textwidth]{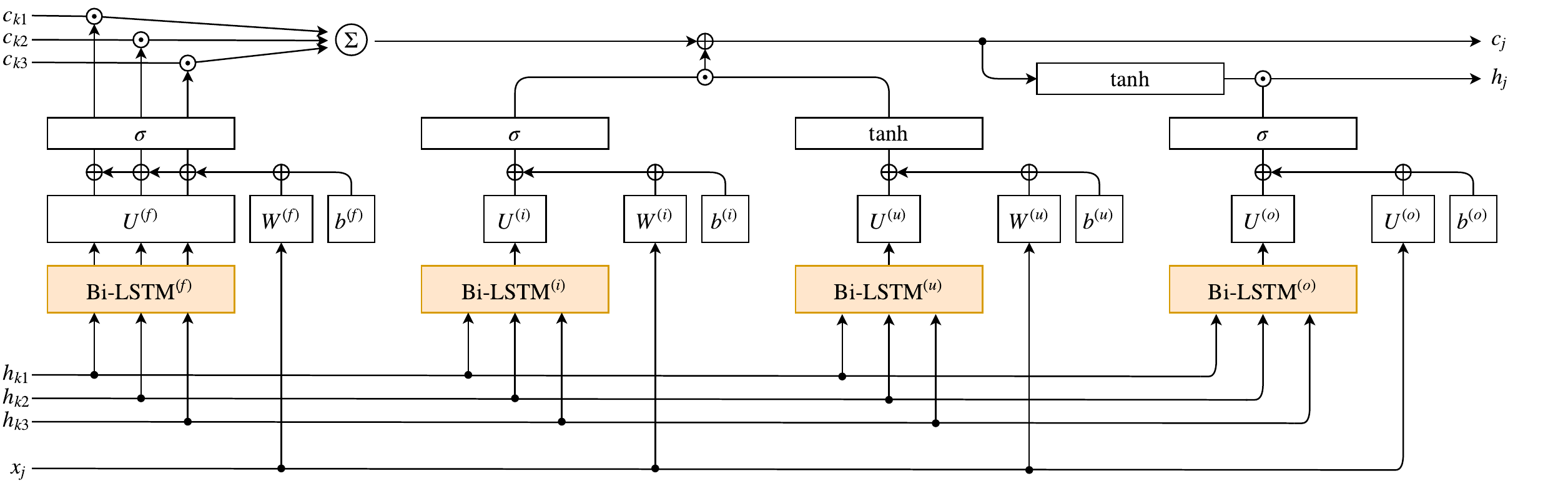}}
\caption{Multi-way Tree-LSTM architecture.}
\label{shidolstm}
\end{figure*}

In Multi-way Tree-LSTM, we add an ordinary chain-like LSTM to each gate immediately before linear transformation $U^{(\ast)}$ to flexibly adapt to a node that has an arbitrary number of ordered children, as shown in Figure \ref{shidolstm}.
The memory cell $c_t$ and the hidden state $h_t$ at each time step $t$ are updated as follows:
\begin{eqnarray}{rCl}
    \check{\bm{f}_{j}} &=& \mathcal{L}^{(f)}(h_{j_1},\ldots, h_{j_{n_{j}}}), \label{mwff} \\
    \check{u_j} &=& \mathcal{L}^{(u)}(h_{j_1},\ldots, h_{j_{n_{j}}})_{j_{n_j}}, \label{mwuu} \\
	\check{i_j} &=& \mathcal{L}^{(i)}(h_{j_1},\ldots, h_{j_{n_{j}}})_{j_{n_j}}, \label{mwii} \\
	\check{o_j} &=& \mathcal{L}^{(o)}(h_{j_1},\ldots, h_{j_{n_{j}}})_{j_{n_j}}, \label{mwoo} \\
	f_{j_k} &=& \sigma(W^{(f)} x_j + U^{(f)}(\check{f_{j_k}}) + b^{(f)}), \label{mwf} \\
	i_j &=& \sigma(W^{(i)} x_j + U^{(f)}(\check{i_j}) + b^{(i)}), \label{mwi} \\
	o_j &=& \sigma(W^{(o)} x_j + U^{(o)}(\check{o_j}) + b^{(o)}), \label{mwo} \\
	u_j &=& \tanh (W^{(u)} x_j + U^{(u)}(\check{u_j}) + b^{(u)}), \label{mwu} \\
	c_j &=& \sum_{k \in C(j)}{c_k \odot f_{j_k}} + i_j \odot u_j, \label{mwc} \\
	h_j &=& o_j \odot \tanh(c_j). \label{mwh}
\end{eqnarray}
Here, $\mathcal{L}^{(\ast)}$ in \eqref{mwff} to \eqref{mwoo} denotes standard chain-like LSTMs and $\mathcal{L}^{(\ast)}(\bm{x})$ is the result of giving a sequence $\bm{x}$ of vectors to $\mathcal{L}^{(\ast)}$.
Let us note that $\check{\bm{f}_{j}}$ is a sequence of $n_j$ vectors and $\check{u_j}$, $\check{i_j}$, and $\check{o_j}$ are the last vectors in the sequence of $\mathcal{L}^{(u)}$, $\mathcal{L}^{(i)}$, and $\mathcal{L}^{(o)}$, respectively.
Moreover, we adopt bidirectional LSTMs 
\cite{schuster1997bidirectional}
for $\mathcal{L}$ at each gate to carry the information on forward children to backward children and vice versa.
A bidirectional LSTM internally has two LSTMs for the forward and backward directions.
Given an input sequence $\bm{x}=(x_1,\ldots,x_n)$, a bidirectional LSTM feeds $(x_1,\ldots,x_n)$ and $(x_n,\ldots,x_1)$ to its LSTMs and gets sequences $\bm{y^{(1)}}$ and $\bm{y^{(2)}}$, respectively.
The two sequences are then combined as $\bm{y} = ([y^{(1)}_1; y^{(2)}_1], \ldots, [y^{(1)}_n; y^{(2)}_n])$, where $[y^{(1)}_i;  y^{(2)}_i]$ is the concatenation of $y^{(1)}_i$ and $y^{(2)}_i$.
Thanks to bidirectional LSTMs, our Multi-way Tree-LSTM can utilize interactions among children at each gate.

\subsection{Code Summarization Framework}
\begin{figure*}[tbp]  
    \centerline{\includegraphics[width=.9\textwidth]{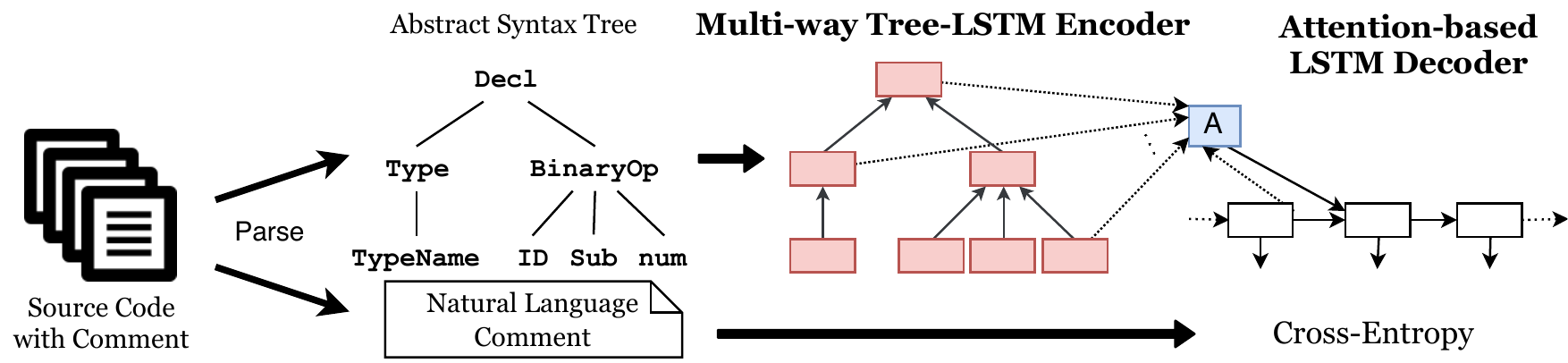}}
    \caption{Overview of our code summarization method.}
    \label{model}
\end{figure*}

An overview of our approach is illustrated in Figure~\ref{model}.
The proposed framework is based on sequence-to-sequence (seq2seq) models \cite{cho2014learning,sutskever2014sequence} and can be roughly divided into three parts: parsing to ASTs, encoding ASTs, and decoding to sequences with attention.
First, we convert each source code into an AST with a standard AST parser.
In our model, each node in the parsed AST is embedded into a vector of fixed dimension. 
The AST with vector-labeled nodes is then encoded by our Multi-way Tree-LSTM.
Finally, the encoded vectors are decoded to a natural language sentence using a LSTM decoder with attention.


\subsubsection{Encoder}
Given an AST, the encoder learns distributed representations of the nodes.
At each node $j$ in the AST, the Multi-way Tree-LSTM encoder $f$ computes the hidden state $h_j^{(e)}$ from input AST node $x_j$ and the hidden states of children $\bm{h}^{(e)}_{C(j)} = \{h_k \mid k \in C(j)\}$ as
\begin{equation*}
    h_j^{(e)} = f(x_j, \bm{h}^{(e)}_{C(j)}).
\end{equation*}

\subsubsection{Attention Mechanism}
The attention mechanism \cite{bahdanau2014} allows neural networks to focus on the relevant part of the input rather than the unrelated part.
This mechanism particularly evolves neural machine translation models \cite{bahdanau2014,luong2015effective,Vaswani2017}. 

In our model, for the hidden state $h^{(e)}_j$ in the encoder at node $j$ and that $h^{(d)}_t$ in the decoder hidden state at time step $t$, the context vector $v_t$ is computed as
\begin{equation*}
    v_t = \sum\limits_{j} \alpha_{tj} \cdot h^{(e)}_j, \label{context}
\end{equation*}
where $\alpha_{tj}$ is the weight between $h^{(e)}_j$ and $h^{(d)}_t$ defined as
\begin{equation*}
    \alpha_{tj} = \frac{\exp(\text{score}(h^{(d)}_t, h^{(e)}_j))}{\sum_j \exp(\text{score}(h^{(d)}_t, h^{(e)}_j))}. \label{alignment}
\end{equation*}
Here, $\text{score}$ is a function that measures the relevance between $h^{(d)}_t$ and $h^{(e)}_j$.
We adopt the simple additive attention \cite{bahdanau2014} as
\begin{equation*}
    \text{score}(h^{(d)}_t, h^{(e)}_j) = w^{(a)\mathrm{T}}\tanh{W^{(a)}[h^{(d)}_t; h^{(e)}_j]},
\end{equation*}
where $w^{(a)}$ and $W^{(a)}$ are model parameters in the attention mechanism.

\begin{figure}[tbp]
\centerline{\includegraphics[width=0.7\textwidth]{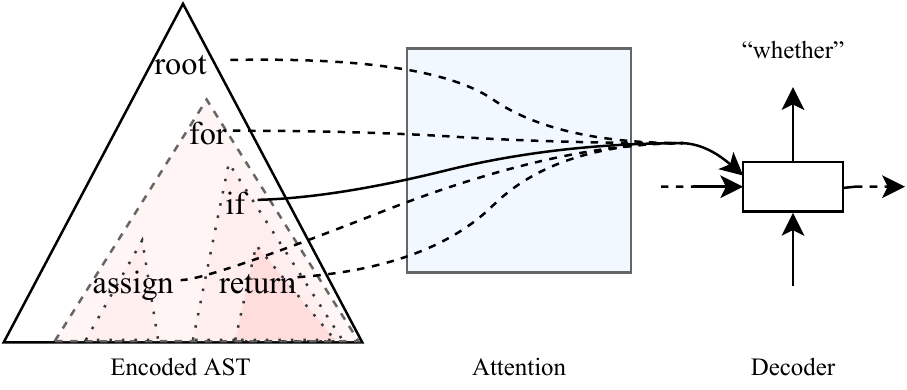}}
\caption{
    Attention mechanism focuses on subtrees of an AST.
    The decoder can focus on important subtrees for summarizing source code.
}
\label{subtree}
\end{figure}

The attention mechanism works between source code and natural language as well.
For example, the token ``='' can be translated directly into ``equal''.
Moreover, in our model, the attention mechanism can focus on {\em subtrees of an AST} as shown in Figure \ref{subtree}.
In ASTs, subtrees are meaningful components in source code such as single expressions, ``if'' statements, and loop statements.
It is possible to focus on such components of various sizes by using the attention mechanism at each node in the tree-structured encoder.

\subsubsection{Decoder}
The decoder decodes the hidden states $h^{(e)}_1, \ldots, h^{(e)}_n$ in the encoder
to a sentence $\bm{y} = (y_1, \ldots , y_m)$ in the target language.
Following \cite{bahdanau2014}, at time step $t$, the LSTM decoder $g$ computes the hidden state $h^{(d)}_t$ as
\begin{equation*}
    h^{(d)}_t = g([y_{t-1}; v_{t-1}], h^{(d)}_{t-1}).
\end{equation*}
Finally, the hidden state $h^{(d)}_t$ in the decoder is projected to a word $y_t$ as $p(y_t)=\text{softmax}(W^{(p)}h^{(d)}_t + b^{(p)})$, where $p(y_t)$ is the predicted probability distribution of $y_t$, and $W^{(p)}$, $b^{(p)}$ are model parameters of the projection layer.
The model parameters are trained by minimizing cross entropy expressed as
\begin{equation*}
    - \sum_t q(y_t)\log p(y_t),
\end{equation*}
where $q(y_t)$ is the true distribution of the word $y_t$.

\section{Experiments}\label{sec:expe}
We conducted comparative experiments with the above framework.
In order to fairly compare the ability of encoders, we made all parts other than those the same as much as possible.

\subsection{Dataset}\label{dataset}
We performed computational experiments with a dataset consisting of pairs of a method written in Java and a Javadoc documentation comment collected by \cite{Hu2018a}.
Since comments are not always given in an appropriate manner, we filtered pairs with comments with one-word descriptions, constructors, setters, getters, and tester methods from the dataset, as in Hu \textit{et al} \cite{Hu2018a}.
Moreover, when a comment has two or more sentences, we only used the first sentence since it typically expresses the functionality of the method.
Hu \textit{et al} truncated the encoded sequences obtained from the dataset to some fixed length.
However, similar truncation cannot be applied directly to ASTs. Therefore, we only use ASTs with nodes at most 100.
Finally, we used $243,183$ samples for training, $29,155$ for validation, and $33,010$ for testing.
Likewise many NLP studies, we limited the vocabulary of identifiers to $30,000$ and those exceeding the limit were replaced with a special token, $\langle$UNK-ID$\rangle$.
We also limited the vocabulary of literals to $1,000$, with the remaining string literals and number literals replaced with $\langle$UNK-STR$\rangle$ and $\langle$UNK-NUM$\rangle$, respectively.

\subsection{Baselines}
In addition to the CODE-NN and DeepCom mentioned in Section~\ref{sec:related}, we compared our model with the Transformer model \cite{Vaswani2017}, a state-of-the-art natural language translation model consisting of only attention mechanisms for both the encoder and the decoder, and attention-based seq2seq models using Child-sum Tree-LSTM and N-ary Tree-LSTM \cite{tai2015} as the encoder.
Although the Transformer, was not designed for the purpose of source code summarization, we include this approach in the experiment\footnote{In the transformer model, we use the same decoder as \cite{Vaswani2017}. Unlike \cite{Vaswani2017}, we set the number of layers to 3 (originally 6) and the dimensions of embedding and model parameters to 256 (originally 512).}.
In the N-ary Tree-LSTM model, it is difficult to use ASTs as input since they have an arbitrary number of children.
Therefore, we converted ASTs into binary trees with a standard binarization technique.
Features of each model used in our experiments are shown in Table \ref{characteristics}.
\begin{table}[!t]
	\caption{
	    Feature comparison of previous and our methods.
	    ``Token order'' indicates whether the order of input tokens can be recovered from the input of the encoders.
	    ``Children order'' indicates whether the order of children in ASTs can be recovered.
	    ``Attention to subtrees'' indicates whether the attention mechanisms in their methods can properly focus on subtrees in ASTs.
	}
	\renewcommand{\arraystretch}{1.3}
	\label{characteristics}
	\centering
	\begin{tabular}{|c||c|c|c|c|}
	    \hline
		Approaches & \begin{tabular}{{@{}c@{}}}Input\\ format\end{tabular} & \begin{tabular}{{@{}c@{}}}Token\\ order\end{tabular} & \begin{tabular}{{@{}c@{}}}Children\\ order\end{tabular} & \begin{tabular}{{@{}c@{}}}Attention\\ to subtrees\end{tabular} \\ \hline \hline
		CODE-NN & Set & No & - & -   \\
		DeepCom & Sequence & Yes & Yes & No   \\
		Multi-way (Ours) & Tree & Yes & Yes & Yes \\ \hline
		Child-sum & Tree & No & No & Yes \\
		N-ary & Tree* & Yes & Yes & Yes \\ \hline
		Transformer & Sequence & Yes & - & -\\
	    \hline
	\end{tabular}
	\begin{flushleft}
	*\footnotesize{Note: N-ary Tree-LSTMs can handle trees whose node has fixed number of children only.}
	\end{flushleft}
\end{table}
The attention mechanism in DeepCom cannot focus on subtrees in the AST since, with their sequence encoding scheme from the AST, the attention mechanism can focus on only prefixes of the encoded sequence, which do not correspond to subtrees of the AST.
In contrast to DeepCom, our proposed model can focus on subtrees in the AST. Subtrees form ``chunks of meaning'' in a method. This may be useful to translate a method into a natural language sentence.

\subsection{Implementation}
Using the dataset described in Section~\ref{dataset}, we trained the models, validated them after every epoch, and tested them.
The models\footnote{Codes are available at \url{https://github.com/sh1doy/summarization_tf}.} were written in TensorFlow and trained on a single GPU (NVIDIA Tesla P100) with the following settings:
\begin{enumerate}
 \item We used a mini-batch size of 80 in training.
 \item The adaptive moment estimation (Adam) algorithm \cite{kingma2014adam} was used with the learning rate set to 0.001 for optimization.
 \item Both encoders and decoders were two-layered with shortcut connections~\cite{he2016deep}
 .
 \item We also implemented a one-layered encoder for Muti-way Tree-LSTM.
 \item Word embedding and hidden states of the encoder and decoder were all 256 dimensional.
 \item To avoid overfitting, we adopted dropout
 \cite{srivastava2014dropout}
 with a drop probability of 0.5.
\end{enumerate}

\subsection{Evaluation Metrics} 
We evaluated the models in several metrics covering different contexts.
BLEU (BLEU-N) \cite{papineni2002bleu} is a metric evaluating N-gram overlaps between two sentences.
CIDEr \cite{vedantam2015cider} is a consensus-based metric for evaluating image captioning.
METEOR \cite{banerjee2005meteor} is a metric based on the weighted mean of the unigram precision and recall.
RIBES \cite{isozaki2010automatic} is a metric based on rank correlation coefficients with word precision.
ROUGE-L \cite{lin2004rouge} is a metric for summaries and is based on the longest common subsequence between two summaries.

\section{Results}\label{sec:result}
In this section, we provide and answer the following two research questions:
\begin{enumerate}
    \item[RQ1] {\em How effective is the Multi-way Tree-LSTM in source code summarization compared with some baseline approaches and the two conventional Tree-LSTMs?}
    \item[RQ2] {\em How effective are our and other methods when varying comment lengths, AST sizes, and the maximum number of children?}
\end{enumerate}

\subsection{Experimental Analysis}

\begin{figure}[htbp]
    \centering
    \subfloat[BLEU-4 score for various comment lengths \label{bleuvscomment}]{%
        \includegraphics[width=0.7\textwidth]{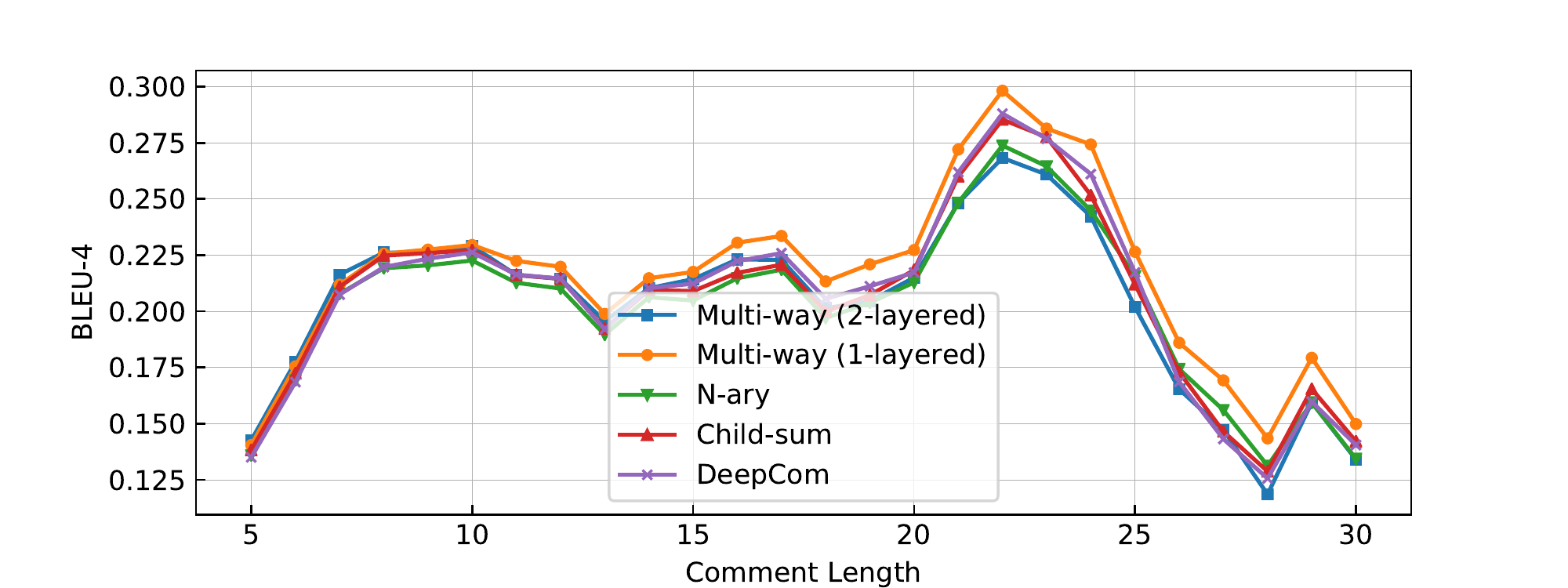}}
    \\
    \subfloat[BLEU-4 score for various numbers of nodes in ASTs
    \label{bleuvscode}]{
        \includegraphics[width=0.7\textwidth]{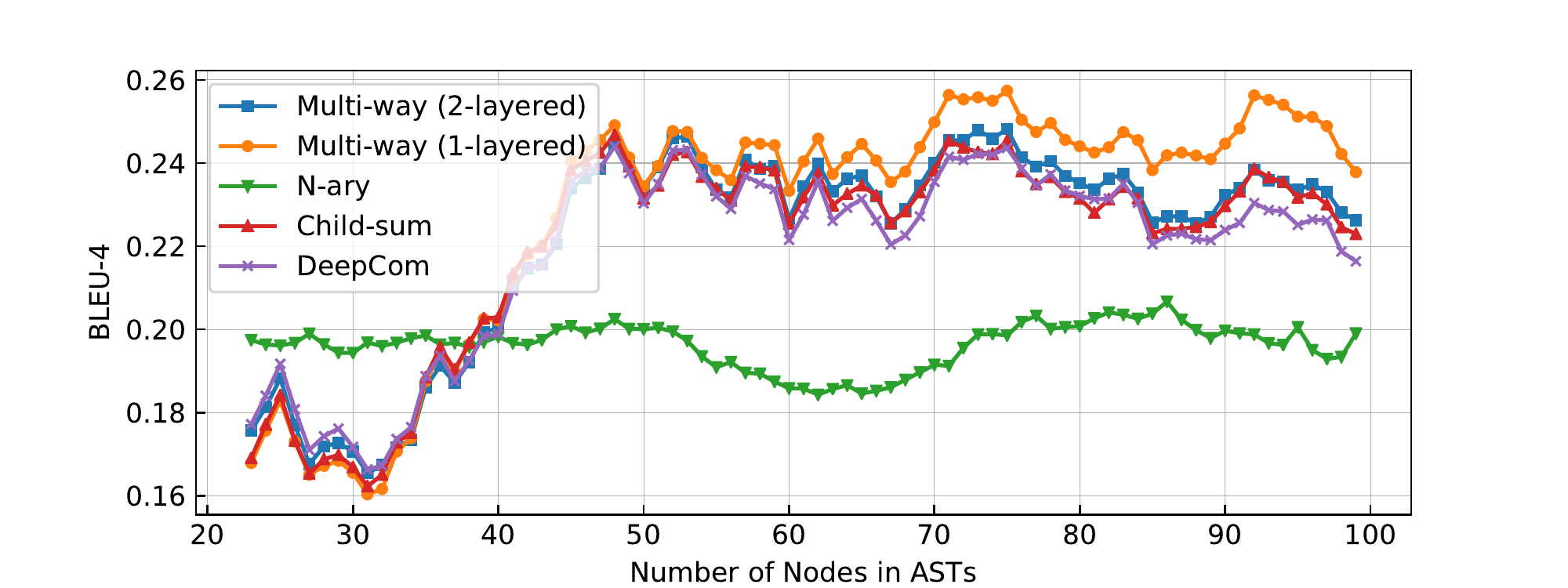}}
    \\
    \subfloat[BLEU-4 score for the maximum degree of ASTs
    \label{bleuvschildren}]{
        \includegraphics[width=0.7\textwidth]{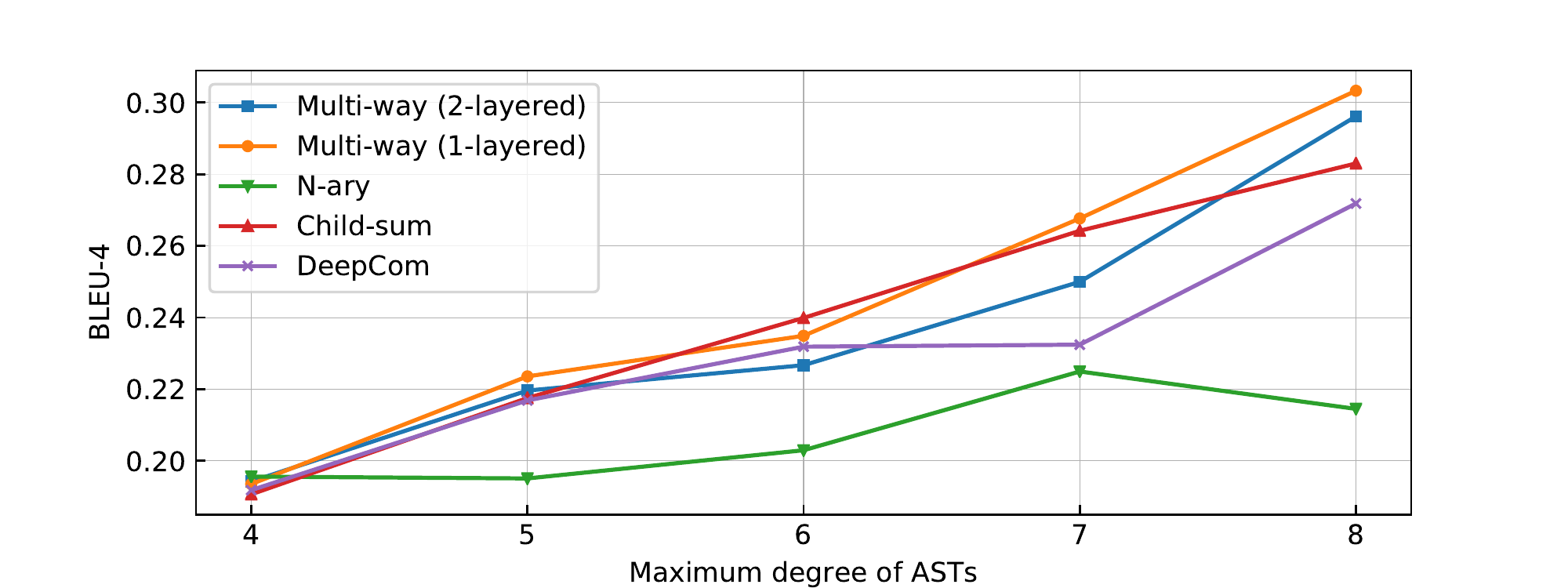}}
\caption{Comparison of BLEU-4 scores with other models by varying (a) the lengths of comments (the number of words in comments), (b) the numbers of nodes in ASTs, and (c) the maximum degrees of ASTs.
}
\label{length}
\end{figure}

The detail of the experimental results are shown in Table \ref{result} and Figure~\ref{length}.
TABLE~\ref{result} shows the comparison among our and other methods in several evaluation criteria. 
\begin{table}[t]
	\centering
	\caption{Evaluation of summaries generated by our and other methods.}
	\label{result}
	\subfloat[BLEU scores\label{bleutable1}]{
    	\begin{tabular}{|c||c|c|c|c|}
    	    \hline
    		Approach                & BLEU-1      & BLEU-2      & BLEU-3      & BLEU-4      \\ \hline \hline
    		CODE-NN                 & 0.2779      & 0.2208      & 0.1974      & 0.1829      \\
    		DeepCom                 & 0.2894      & 0.2361      & 0.2132      & 0.1988      \\
    		Multi-way (1-layered)   & 0.2968      & \bf{0.2431} & \bf{0.2191} & \bf{0.2040} \\
            Multi-way (2-layered)   & \bf{0.2984} & 0.2413      & 0.2170      & 0.2015      \\
    		\hline     
    		Child-sum               & 0.2968      & 0.2400      & 0.2153      & 0.1997      \\
    		N-ary                   & 0.2944      & 0.2366      & 0.2117      & 0.1959      \\ \hline
    		Transformer             & 0.1173      & 0.0583      & 0.0378      & 0.0249      \\
    		\hline
		\end{tabular}
	}
	\\
	\subfloat[Other metrics\label{bleutable2}]{
    	\begin{tabular}{|c||c|c|c|c|c|}
    	    \hline
    		Approach                & CIDEr       & METEOR      & RIBES       & ROUGE       \\ \hline \hline
    		CODE-NN                 & 1.8170      & 0.1643      & 0.2571      & 0.3051      \\
    		DeepCom                 & 1.9753      & 0.1608      & 0.2727      & 0.3130      \\
    		Multi-way (1-layered)   & \bf{2.0196} & 0.1665      & 0.2825      & 0.3221      \\
            Multi-way (2-layered)   & 1.9886      & \bf{0.1752} & 0.2814      & \bf{0.3259} \\
    		\hline     
    		Child-sum               & 1.9693      & 0.1701      & \bf{0.2844} & 0.3238      \\
    		N-ary                   & 1.9327      & 0.1699      & 0.2802      & 0.3223      \\ \hline
    		Transformer             & 0.2102      & 0.0744      & 0.0967      & 0.1556      \\
    		\hline
    	\end{tabular}
	}
\end{table}

For RQ1, our methods (1-layered) and (2-layered) are better than the previous methods CODE-NN and DeepCom in all evaluation criteria. Moreover, the conventional Tree-LSTMs are even better than the previous methods.
In consequence of these facts, we can see that ASTs should be treated as they are without encoding to sequences, and Tree-LSTMs, including our proposal, can leverage the tree-structured nature of ASTs.
The experiment also shows that Transformer, which is one of the state-of-the-art methods in neural machine translation, does not work well for source code summarization.
This suggests that source code is quite different from natural language sentences.
It would be interesting that our one-layered Multi-way Tree-LSTM encoder model outperforms its two-layered model in multiple evaluation criteria, whereas multi-layered seq2seq models are better than single-layered.
For example, the BLEU-4 score of a one-layered seq2seq encoder model (one-layered version of DeepCom) was $0.1895$ in our implementation, which is lower than the two-layered one (original DeepCom).

Figure~\ref{length} shows the detail of BLEU-4 scores one some methods based on ASTs when varying comment lengths (Figure~\ref{bleuvscomment}), AST sizes (Figure~\ref{bleuvscode}) and maximum children sizes (Figure~\ref{bleuvschildren}).
In the following, we call the maximum number of children the maximum degree of the AST.

For RQ2, we can conclude that our summarization models based on Multi-way Tree-LSTMs are better than other models.
Although we do not see any considerable difference among our and other models in various comment lengths (Figure \ref{bleuvscomment}), our one-layered model is still better than other models when generating summaries of moderate lengths.
On the other hand, AST sizes and their maximum degrees have an impact on the quality of summaries. 
We find that our one-layered Multi-way Tree-LSTM model significantly outperforms the other models when ASTs have many nodes or large degree as shown in Figures~\ref{bleuvscode} and \ref{bleuvschildren}.
It is worth noting that ASTs containing many nodes are needed to be appropriately commented, and hence we would like to say that our model is more suitable for practical purposes.

\subsection{Output Examples}
Table~\ref{example} shows some examples of summaries generated by our method.
We only picked some interesting examples and hence do not claim that our method always generates such a summary.
The summaries are quite natural compared with the original documentation comments in the dataset.
In some cases, our model generated exactly the same sentences as in (1), (2).
In other cases, our model expressed almost the same meaning in different words (3), (4).
It is worth noting that some summaries are more expressive than the original sentences as in (5), (6).
\begin{table*}[t]
    \centering
    \caption{Output examples.}
    \label{example}
    \begin{tabular}{|c|c|l|}
        \hline
        ID & \multicolumn{2}{|c|}{Source code and comment} \\
        \hline \hline
        \multirow{4}{*}{1}
            & Source code & 
                \begin{lstlisting}
public static Charset toCharset(Charset charset){
    return charset == null ? Charset.defaultCharset(): charset;
}
                \end{lstlisting}
            \\ \cline{2-3}
            & Gold           & \begin{tabular}{{@{}c@{}}}
                Returns the given Charset or the default Charset if the given Charset is null
            \end{tabular} \\ \cline{2-3}
            & Generated      & \begin{tabular}{{@{}c@{}}}
                Returns the given Charset or the default Charset if the given Charset is null
            \end{tabular} \\ 
            \hline 

        \multirow{4}{*}{2}
            & Source code & 
                \begin{lstlisting}
public boolean more() throws JSONException {
  next();
  if (end()) {
    return false;
  }
  back();
  return true;
}
                \end{lstlisting}
            \\ \cline{2-3}
            & Gold           & \begin{tabular}{{@{}c@{}}}
                Determine if the source string still contains characters that next() can consume
            \end{tabular} \\ \cline{2-3}
            & Generated      & \begin{tabular}{{@{}c@{}}}
                Determine if the source string still contains characters that next() can consume
            \end{tabular} \\ 
            \hline 

        \multirow{4}{*}{3}
            & Source code & 
                \begin{lstlisting}
@JsonIgnore public boolean isDeleted(){
  return state.equals(Experiment.State.DELETED);
}
                \end{lstlisting}
            \\ \cline{2-3}
            & Gold           & \begin{tabular}{{@{}c@{}}}
                Signals if this experiment is deleted
            \end{tabular} \\ \cline{2-3}
            & Generated      & \begin{tabular}{{@{}c@{}}}
                Returns true if this session has been deleted
            \end{tabular} \\ 
            \hline 

        \multirow{4}{*}{4}
            & Source code & 
                \begin{lstlisting}
public static boolean isEmpty(CharSequence str){
  return TextUtils.isEmpty(str);
}
                \end{lstlisting}
            \\ \cline{2-3}
            & Gold           & \begin{tabular}{{@{}c@{}}}
                Check if a string is empty
            \end{tabular} \\ \cline{2-3}
            & Generated      & \begin{tabular}{{@{}c@{}}}
                Returns true if the string is null or 0-length
            \end{tabular} \\ 
            \hline 

        \multirow{4}{*}{5}
            & Source code & 
                \begin{lstlisting}
public static boolean removeFile(File file){
  if (fileExists(file)) {
    return file.delete();
  }
 else {
    return true;
  }
}
                \end{lstlisting}
            \\ \cline{2-3}
            & Gold           & \begin{tabular}{{@{}c@{}}}
                Remove a file
            \end{tabular} \\ \cline{2-3}
            & Generated      & \begin{tabular}{{@{}c@{}}}
                Delete a file or directory if it exists
            \end{tabular} \\ 
            \hline 

        \multirow{4}{*}{6}
            & Source code & 
                \begin{lstlisting}
public void dismissProgressDialog(){
  if (isProgressDialogShowing()) {
    mProgressDialog.dismiss();
    mProgressDialog=null;
  }
}
                \end{lstlisting}
            \\ \cline{2-3}
            & Gold           & \begin{tabular}{{@{}c@{}}}
                Dismiss progress dialog
            \end{tabular} \\ \cline{2-3}
            & Generated      & \begin{tabular}{{@{}c@{}}}
                Hide the progress dialog if it is visible
            \end{tabular} \\ 
            \hline

    \end{tabular}
\end{table*}

\section{Conclusion}\label{sec:conc}

Neural network approaches are certainly successful in machine translation. These approaches are expected to be so in source code summarization since we can see it as translations from source code to natural language sentences. However, there is an indispensable difference between source code and natural language: Source code is essentially structured. 
This fact arises a natural question: How do we use structural information in neural networks?
Fortunately, the essential structure of source code forms a tree, namely an AST. 
This suggests that neural networks for trees would be useful in source code summarization. 

In this paper, we proposed an extension of Tree-LSTM on the basis of the work of Tai {\it et al.} \cite{tai2015}, which is a generalization of LSTM for trees.
Our extension obtains a distributed representations of ordered trees, such as ASTs, which cannot be directly handled by the known Tree-LSTMs since they have an arbitrary number of ordered children.
We applied our extension to source code summarization as the encoder and compared with other baseline methods. Our experimental results show that our extension is suitable for dealing with ASTs, and code summarization framework with our extension can generate high-quality summaries.
We would like to mention that some summaries generated by our method are more expressive than the original handmade summaries. This indicates the effectiveness of automatic document generation with neural networks for ASTs.

\bibliographystyle{plain}
\bibliography{paper}

\end{document}